# Comparing Object Recognition in Humans and Deep Convolutional Neural Networks - An Eye Tracking Study

Leonard Elia van Dyck[1,2*], Roland Kwitt[3], Sebastian Jochen Denzler[1], & Walter Roland Gruber[1,2]

[1]*Department of Psychology, University of Salzburg, Austria*
[2]*Center for Cognitive Neuroscience, University of Salzburg, Austria*
[3]*Department of Computer Science, University of Salzburg, Austria*
*To whom correspondence should be addressed:* leonard.vandyck@plus.ac.at

## Abstract

Deep convolutional neural networks (DCNNs) and the ventral visual pathway share vast architectural and functional similarities in visual challenges such as object recognition. Recent insights have demonstrated that both hierarchical cascades can be compared in terms of both exerted behavior and underlying activation. However, these approaches ignore key differences in spatial priorities of information processing. In this proof-of-concept study, we demonstrate a comparison of human observers (N = 45) and three feedforward DCNNs through eye tracking and saliency maps. The results reveal fundamentally different resolutions in both visualization methods that need to be considered for an insightful comparison. Moreover, we provide evidence that a DCNN with biologically plausible receptive field sizes called *vNet* reveals higher agreement with human viewing behavior as contrasted with a standard ResNet architecture. We find that image-specific factors such as category, animacy, arousal, and valence have a direct link to the agreement of spatial object recognition priorities in humans and DCNNs, while other measures such as difficulty and general image properties do not. With this approach, we try to open up new perspectives at the intersection of biological and computer vision research.

# 1 Introduction

In the last few years, advances in deep learning have turned rather simple convolutional neural networks, once developed to simulate the complex nature of biological vision, into sophisticated objects of investigation themselves. Especially the increasing synergy between neural and computer sciences has facilitated this interdisciplinary progress with the aim to enable machines to see and to further the understanding of visual perception in living organisms along the way. Computer vision has given rise to deep convolutional neural networks (DCNNs) that exceed human benchmark performance in key challenges of visual perception (He, Zhang, Ren, & Sun, 2015; Krizhevsky, Sutskever, & Hinton, 2012). Among them, the fundamental ability of core object recognition, which allows humans to identify an enormous number of objects despite their substantial variations in appearance (DiCarlo, Zoccolan, & Rust, 2012) and thus to classify visual inputs into meaningful categories based on previously acquired knowledge (Cadieu et al., 2014).

In the brain, this information processing task is solved particularly by the ventral visual pathway (Ishai, Ungerleider, Martin, Schouten, & Haxby, 1999), which passes information through a hierarchical cascade of retinal ganglion cells (RGC), lateral geniculate nucleus (LGN), visual cortex areas (V1, V2, and V4), and inferior temporal cortex (ITC) (DiCarlo et al., 2012; Riesenhuber & Poggio, 1999; Rolls, 2000; Tanaka, 1996). This organization shares vast similarities with the purely feedforward architectures of DCNNs in a way that visual information can pass through by means of a single end-to-end sweep (see Figure 1). While in most cases this processing mechanism seems to suffice for so called *early-solved* natural images, a substantial body of literature proposes that especially *late-solved* challenge images benefit from recurrent processing through neural interconnections and loops (Kar & DiCarlo, 2020; Kar, Kubilius, Schmidt, Issa, & DiCarlo, 2019; Lamme & Roelfsema, 2000). Moreover, electrophysiological findings therefore suggest that recurrence may set in increasingly after around 150ms to stimulus onset (Cichy, Pantazis, & Oliva, 2014; Contini, Wardle, & Carlson, 2017; DiCarlo & Cox, 2007; Rajaei, Mohsenzadeh, Ebrahimpour, & Khaligh-Razavi, 2019; Seijdel et al., 2020; Tang et al., 2018). Interestingly, DCNNs seem to face difficulties in recognizing exactly these late-solved (Kar et al., 2019) and manipulated challenge images (Dodge & Karam, 2017; Geirhos et al., 2017; Geirhos, Temme, et al., 2018; van Dyck & Gruber, 2020), which may require additional recurrent processing.

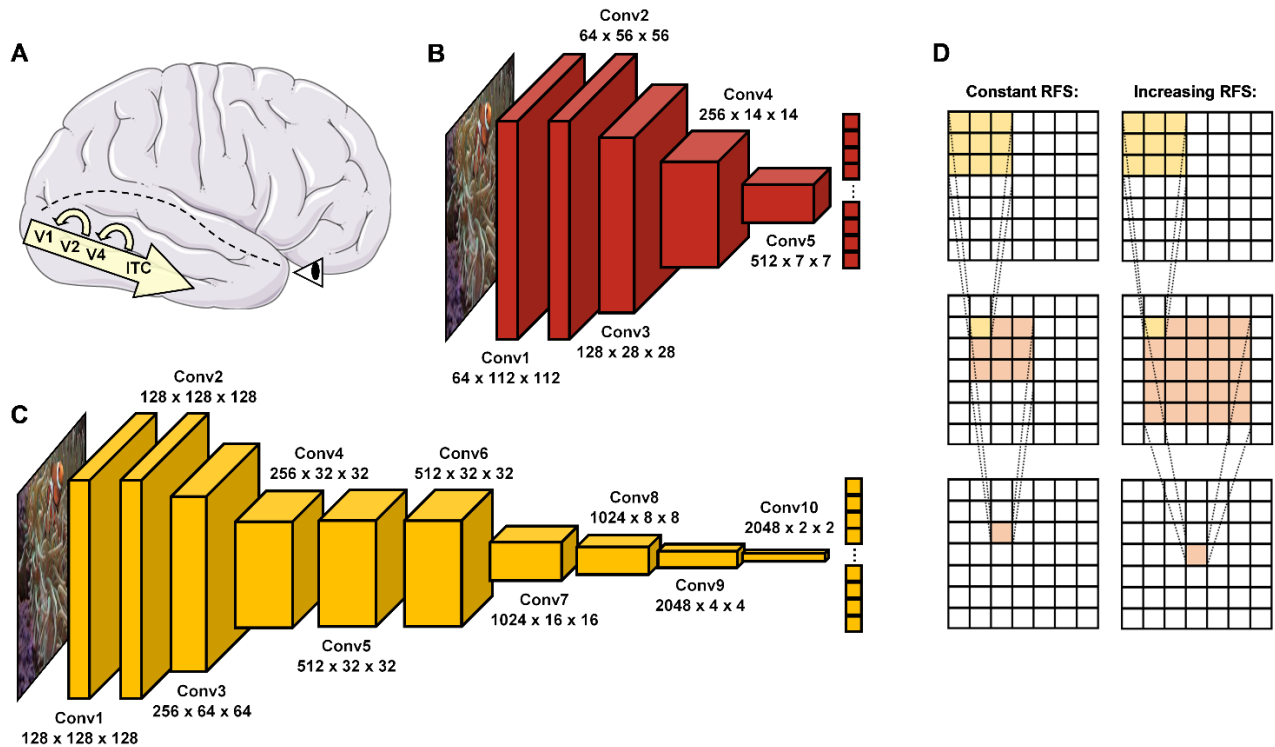

**Figure 1.** Object recognition in the human brain and DCNNs. (**A**) In the brain, visual information enters via the retina before it passes through the ventral visual pathway, consisting of the visual cortex areas (V1, V2, and V4) and inferior temporal cortex (ITC). After a first feedforward sweep of information (~ 150ms), recurrent processes reconnecting higher to lower areas of this hierarchical cascade become activated and allow more in-depth visual processing. (**B**) ResNet18 (He, Zhang, Ren, & Sun, 2016) is a standard DCNN with 5 convolutional layers. (**C**) vNet (Mehrer, Spoerer, Jones, Kriegeskorte, & Kietzmann, 2021) is a novel DCNN with 10 convolutional layers and modified effective kernel sizes, which simulate the progressively increasing *receptive field size* (RFS) in the ventral visual pathway. (**D**) Schematic overview of convolutions with a constant and increasing RFS. An increasing RFS raises the number of pixels represented within individual neurons of the later layers. **Note:** Below individual layer names, the first number represents the feature map size while the second and third indicate the obtained output size.

Naturally, images of objects, regardless of whether represented on a biological retina or in a computer matrix, are highly complex. Therefore, in the brain, object recognition is influenced by a number of bottom-up (Rutishauser, Walther, Koch, & Perona, 2004) and top-down processes (Bar, 2003; Bar et al., 2006). In simplified terms however, it is thought to solve this challenge by transferring given two-dimensional information into an invariant three-dimensional representation, encoded in an even higher-dimensional neuronal space (DiCarlo et al., 2012). As pointed out by David Marr (1982), the two-dimensional image allows first and foremost the extraction of shape features such as edges and regions, the precursors of the so-called *primal sketch* (ibid., p. 37). Then, when textures and shades start to enrich the outline, a *2.5D sketch* (ibid., p. 37) emerges. Finally, in combination with previously acquired knowledge, the representation of an invariant *3D model* can be inferred (ibid., p. 37). These processing steps are reflected not only by the architecture of the ventral visual pathway but can also be found in its artificial replica. While in DCNNs, earlier layers are mostly sensitive to specific configurations of edges, blobs, and colors (e.g., the edge of an orange stripe), the following convolutions start to combine them into texture-like feature groups (e.g., an orange-white striped pattern), until later layers assemble whole object parts (e.g., the fin of an anemone fish) and eventually infer the object class label (e.g., an anemone fish). Interestingly, recent literature suggests that DCNNs trained on the ImageNet dataset (Deng et al., 2009; Russakovsky et al., 2015) are strongly biased towards texture and use it more frequently rather than shape information to classify images (Geirhos, Rubisch, et al., 2018). This preference seems to contradict findings in humans that clearly identify shape as the single most important cue for object recognition (Landau, Smith, & Jones, 1988). In addition, several studies have examined the effect of context in human object recognition (Greene & Oliva, 2009; Oliva & Torralba, 2007). Contextual cues are representational regularities such as spatial layout (e.g., houses are usually located on the ground), inter-object dependencies (e.g., houses are usually attached to a street), point of view (e.g., houses are usually looked at from a specific perspective), and other summary statistics (i.e., entropy and power spectral density). Likewise, these same cues are certainly not meaningless to DCNNs. In fact, mainly work around removing or manipulating one of these hints such as the object's regular pose (Alcorn et al., 2019) or background (Beery, Van Horn, & Perona, 2018) have demonstrated that indeed contextual learning, which is common practice in learning systems, can be found here as well (Geirhos et al., 2020).

Nevertheless, a key difference between humans and DCNNs lies within the modulating effects of appraisal. While millions of years of evolution have tuned *in vivo* object recognizers such as humans to seek and avoid different kinds of stimuli (e.g., find nourishment and avoid dangers quickly), this subjective experience of for example arousal and valence is lacking completely in *in silico* models. Furthermore, the neuroscientific literature contains many examples suggesting that particularly the automatic detection of fear-relevant and threatful objects is solved by an even faster subcortical route, which skips parts of the ventral visual pathway through shortcuts to the amygdala (Öhman, 2005; Pessoa & Adolphs, 2010). This for example might enable threat-superiority effects in terms of reaction times and reduced position effects in a visual search paradigm (Blanchette, 2006). While the plausibility of this so-called *low road* is highly discussed (Cauchoix & Crouzet, 2013), the differences in performance are well-documented.

As an interim summary, it can be noted that the ventral visual pathway and DCNNs suggest conceptual overlaps but also substantial differences in many regards. Hence, more recent evidence from Mehrer et al. (2021) further highlights the importance of biological plausibility for the fit between brain and DCNN activity, as their novel architecture called *vNet* simulates the progressively increasing foveal *receptive field size* (hereafter abbreviated as RFS) along the ventral visual pathway (Grill-Spector, Weiner, Kay, & Gomez, 2017; Wandell & Winawer, 2015). However, as their analyses point out, this modification does not lead to higher congruence with for example fMRI activity of human observers when compared to a standard architecture such as AlexNet (Krizhevsky et al., 2012). This raises many questions about the definite impact of this RFS modification. Therefore, as vNet's hierarchical organization is designed to resemble that of the ventral visual pathway more accurately as compared to a standard DCNN, here we hypothesize that the major advantage of vNet may not be visible within more brain-like activations, as also not found by Mehrer et al. (2021), but rather more similar spatial priorities of information processing compared through eye tracking and GradCAM. Consequently, following an important distinction in human-machine comparisons by Firestone (2020), we believe that this resemblance in underlying *competence* should lead to higher similarity in object recognition behavior and further observable *performance*.

As in this specific human-machine comparison rather divergent architectures are compared, we believe that a RFS modification can result in more similar spatial priorities of information processing and likewise object recognition behavior, without immediately suggesting a higher match between activity patterns of neural components and individual DCNN layers.

As DCNNs are getting more and more complex, several *attribution*-tools have been developed to understand (to some extent) their classifications. At first glance, these new visualization algorithms, also called *saliency maps*, resemble well-established methods in cognitive neuroscience. In eye tracking, the execution of a visual task is analyzed by mapping gaze behavior onto specific regions of interest, which receive special cognitive or computational priorities during information processing. While eye tracking and saliency maps share this concept, the methodological way this is achieved seems fundamentally different. In eye tracking measurements, a human observer is presented with a stimulus, which is only presented for a limited time, while viewing behavior is being recorded. Saliency maps such as *Gradient-weighted Class Activation Mapping*, also known as *GradCAM* (Selvaraju et al., 2017), extract class activations within a specific layer of the DCNN to explain obtained predictions. Therefore, the algorithm uses the gradient of the loss function to compute a weight for every feature map. The weighted sum of these activations is class-discriminative and hence allows the localization and visualization of all relevant regions that contributed to the probability of a given class. However, DCNNs do not operate on a meaningful time scale (other than related to computational power) when trying to classify images. Despite the conceptual similarity between eye tracking and saliency maps, only a few attempts have been made to draw this comparison of black boxes (Ebrahimpour, Falandays, Spevack, & Noelle, 2019). Importantly, a major challenge in this field is to encourage and conduct fair human-machine comparisons, as it is only possible to infer similarities and differences if there are no fundamental constraints within the comparison itself (Firestone, 2020; Funke et al., 2020). In this study, we compare human eye tracking to DCNN saliency maps in an approximately species-fair object recognition task and examine a wide range of possible factors influencing similarity measures.

# 2   Methods

## 2.1  General Procedure

In order to test our hypotheses, we designed a fair human-machine comparison that allowed us to investigate behavioral, eye tracking, and physiological data from human observers performing a laboratory experiment, as well as predictions and activations of three DCNN architectures on identical visual stimuli and under roughly similar conditions. The main task in this experiment was to categorize briefly shown images based on a forced-choice format of 12 basic-level categories, namely *human*, *dog*, *cat*, *bird*, *fish*, *snake*, *car*, *train*, *house*, *bed*, *flower*, *ball* (see 4.4 Methods). Basic-level categories (e.g., *dog*) were chosen over detailed category concepts which are more often used in the field of computer vision (e.g., *border collie*), as they are more naturally utilized in human object classification (Rosch, 1999).

## 2.2 Human Observers – Eye Tracking Experiment

A total of 45 valid participants (28 female, 17 male) with an age between 18 and 31 years (M = 22.64, SD = 2.57) were tested in the eye tracking experiment. Participants were required to have normal or corrected-to-normal vision without problems of color perception and other eye diseases. Two participants with contact lenses had to be excluded from further analyses due to insufficient eye tracking precision. The experimental procedure was admitted by the University of Salzburg ethics committee, in line with the declaration of Helsinki and agreed to by participants via written consent before the experiment. Psychology students received accredited participation hours for taking part in the experiment.

The experiment consisted of two main tasks (see Figure 2). Participants had to concentrate on a fixation cross for 500ms until an image appeared at center 12.93 degrees of visual angle away on the left or right side. If, as in one condition, the image was presented for a short duration of 150ms, a visual backward mask (1/f noise) followed for the same duration, and the participant had to classify the presented object based on a forced-choice format of 12 basic-level categories by clicking on the respective class symbol. If, as in the other condition, the image was presented for a long duration of 3000ms, participants had to rate it afterwards in its arousal (1 = *Very low*, 4 = *Neither low nor high*, 7 = *Very high*) and valence (1 = *Very negative*, 4 = *Neutral*, 7 = *Very positive*) on a scale from 1 to 7. As both classification and rating tasks were balanced out in occurrence and previously pseudo-randomized, it was impossible for the observers to differentiate between the two conditions before the short presentation time was exceeded. Based on this central assumption, both conditions should not vary in viewing behavior. Participants were familiarized with the experimental procedure during training trials which were excluded from further analyses. The whole experiment consisted of 420 test trials (210 per condition), took about one hour to complete, and was divided into three blocks with resting breaks in between. In this way, a single participant classified one half of the entire dataset and rated the other half. To obtain categorization and rating results for all images, there were two versions of the experiment with interchanged conditions for both halves.

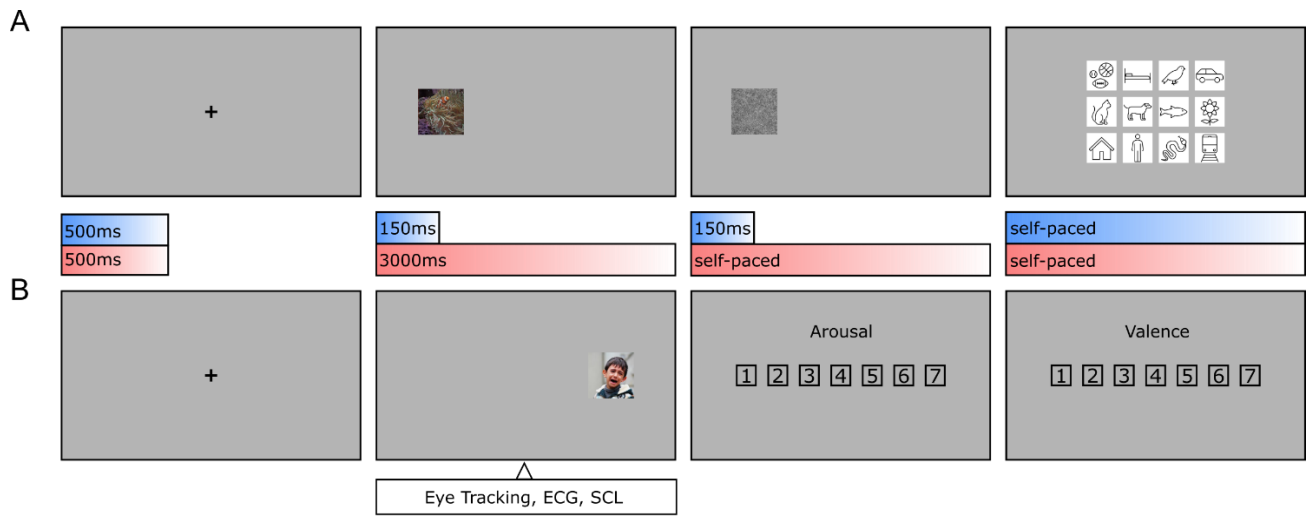

**Figure 2.** Object recognition paradigm. **(A)** During categorization trials, human observers had to focus on a fixation cross (500ms, including a 150ms fixation control) before an image was presented on either the left or the right side (150ms), followed by a visual backward mask (150ms, 1/f noise), and a forced-choice categorization (self-paced, max. 7500ms). **(B)** During rating trials and after the fixation cross, an image was presented again on the left or the right side (3000ms), followed by an arousal rating and valence rating (both self-paced, max. 7500ms each). As conditions were pseudo-randomized, human observers were not able to anticipate, whether an image needed to be categorized or rated until the initial 150ms had elapsed. This allows the assumption that in the first 150ms the conditions should not vary in viewing behavior.

The participants performed the experiment in a laboratory room, where they were seated in front of a screen (1920x1080 pixels, 50Hz) and had to place their head into a chin rest located at a distance of 60cm. The right eye was tracked and recorded with an EyeLink 1000 (SR Research Ltd., Mississauga, ON, Canada) desktop mount, at a sampling rate of 1000Hz. For presentational purposes, original images were scaled by factor two onscreen (448x448) but stayed unchanged in image resolution (224x224). This way, the presented images had 11.52 degrees of visual angle in size. Recorded eye tracking data were preprocessed using DataViewer (Version 4.2, SR Research Ltd., Mississauga, ON, Canada) and analyzed after the participants gaze crossed an invisible boundary framing the entire image. In this way, participants were able to process appearing images already peripherally for the first couple of milliseconds to allow a meaningfully programmed first fixation. Fixations were compiled for the first 150ms within the image. Here, x- and y-coordinates were downscaled again from expanded presentation size (448x448) to original image size (224x224). Average heatmaps were computed from individual sampling points of either the first 150ms (= feedforward) or the entire presentation time after 150ms (= recurrent) within the image using in-house built MATLAB scripts. The obtained heatmaps for all participants were averaged per image, Gaussian filtered with a standard deviation of 15 pixels, and normalized.

Additionally, before the eye tracking experiment started, participants were wired with ECG and SCL electrodes. Physiological measurements were collected using a Varioport biosignal recorder (Becker Meditec, Karlsruhe, Germany) and analyzed with ANSLAB (Blechert, Peyk, Liedlgruber, & Wilhelm, 2016) (Version 2.51, Salzburg, Austria). Here, the relative change in mean activity from a baseline of 1000ms before the image presentation to the time window of 3000ms during the image presentation was used.

**A**  **Dispersion of Fixations along x-Coordinates**

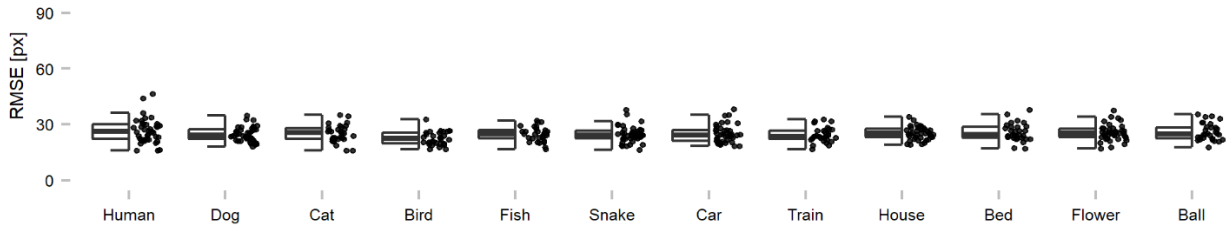

**B**  **... and y-Coordinates**

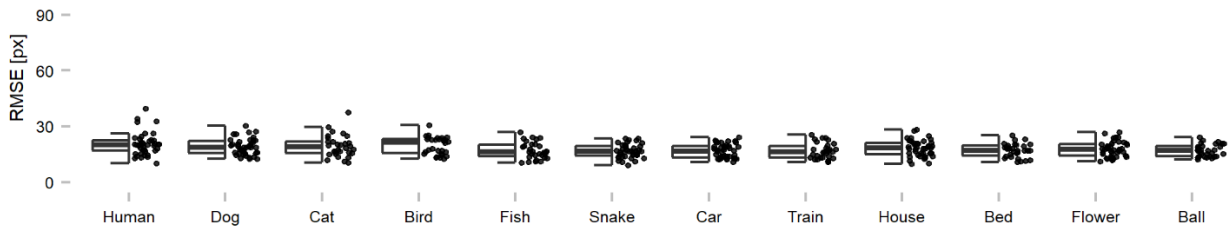

**Figure 3.** Dispersion of eye tracking fixations along x- and y-coordinates across categories. Root Mean Squared Errors (RMSEs) were computed for individual images and indicate the magnitude of deviation among the fixations of individual participants on a single image. Generally, the fixations seemed to be rather precise on an image-by-image level with an average dispersion of below 30 pixels across all participants. This suggests that meaningful features were targeted and that the centroids, which were used for further analyses, can be regarded as characteristic for the human observer sample. The dispersion along (**A**) x-coordinates was slightly higher as compared to (**B**) y-coordinates, which is thought to reflect the reported central fixation and saccadic motor biases. **Note:** The dispersion of an image can also be increased by the presence of multiple meaningful features, without any loss of precision. However, this problem should occur rarely, as most of the images showed only one dominant object.

## 2.3 DCNN Training & GradCAM Saliency Maps

Our implementation (in PyTorch) of vNet follows the original GroupNorm variant of vNet from Mehrer et al. (2021), the ResNet18 architecture follows the original proposal from He et al. (2016). For training both network types, we use the (publicly available) Ecoset *training* set, constrained to the 12 categories described in Section 4.4 below. As the number of images differs significantly across categories, we artificially balance the corpus by drawing (uniformly at random) $N$ images per category, where $N$ corresponds to the number of images in the smallest category (i.e., *fish*). During training, all (~15k) images are first resized to a spatial resolution of 256x256, then cropped to the center square of size 224x224 and eventually normalized (by subtracting the channel-wise mean and dividing by the channel-wise standard deviation, computed from the training corpus). Note that no data augmentation is applied across all experiments. We minimize the cross-entropy loss using stochastic gradient descent (SGD) with momentum (0.9) and weight decay (1e-4) under a cosine learning rate schedule, starting at an initial learning rate of 0.01. We train for 80 epochs using a batch size of 128. Results for the fine-tuned ResNet18 are obtained by replacing the final linear classifier of an ImageNet-trained ResNet18, freezing all earlier layers, and fine-tuning for 50 epochs (in the same setup as described before). When referring to *early*, *middle*, and *late* layers in the manuscript, we refer to GradCAM outputs generated from activations after the 2nd, 6th, and 9th layer for vNet, and activations after the 1st, 3rd, and 4th ResNet18 block (as all ResNet architectures total four blocks). Eventually, if the aim is to investigate the similarity in spatial priorities of information processing that underlie object recognition behavior with the demonstrated approach, the output layer of a DCNN would suffice for the comparison. In this proof-of-concept study, however, we also include earlier layers for sanity checks and further model comparisons.

## 2.4 Dataset

Images were part of 12 basic-level categories from the ecologically motivated Ecoset dataset, which was created by Mehrer et al. (2021) in order to better capture the organization of human-relevant categories. The dataset consisted of 6 animate (namely *human*, *dog*, *cat*, *bird*, *fish*, and *snake*) and 6 inanimate categories (namely *car*, *train*, *house*, *bed*, *flower*, and *ball*) with 30 images per category. During preprocessing, images were randomly drawn from the test set, cropped towards the biggest possible central square, and resized to 224x224 pixels. All images were visually checked and excluded if multiple categories (e.g., *human* and *dog*), overlayed text, or image effects (e.g., greyscale images) were visible or the object was fully removed during preprocessing steps. In categories where less then 30 images from the test set remained, Ecoset images were complemented with ImageNet examples (n = 28, across 4 categories). Additionally, in 3 animate (namely *human*, *dog*, and *snake*) and 3 inanimate categories (namely *car*, *house*, *flower*), 10 images per category of the respective objects were added from the Open Affective Standardized Image Set (OASIS) (Kurdi, Lozano, & Banaji, 2017). Here, arousal and valence ratings from a large number of participants (N = 822) were already available and increased the variability while serving as a sanity check for own ratings. In total, the dataset consisted of 420 test images.

## 3 Results

## 3.1 Performance

The first set of analyses investigated object recognition performance in human observers and DCNNs. Therefore, human predictions obtained during categorization trials (see 4.2 Methods) were compared against model predictions. Generally, as the categorization data were not normally distributed, nonparametric tests were applied to compare the human observer sample against fixed-accuracy values of individual DCNNs. On average, human observers reached a recognition accuracy of 89.96%. One-sample Wilcoxon tests indicated that human observers were significantly outperformed by fine-tuned

ResNet18 with 95.48% (V = 0, CI = [89.52, 91.43], p <.001, r = .85) but significantly more correct than both *trained-from-scratch* ResNet18 (V = 1035, CI = [89.52, 91.43], p <.001, r = .85) and vNet (V = 1034, CI = [89.52, 91.43], p <.001, r = .85) with accuracies of 70.48% and 76.43%, respectively. The results endorse both sides of the literature by demonstrating that especially DCNNs trained on large datasets can exceed human benchmark performance (He et al., 2016; Huang, Liu, Van Der Maaten, & Weinberger, 2017; Krizhevsky et al., 2012; Szegedy et al., 2015), but also simultaneously reminds of possible limits due to the amount of provided training data. Nevertheless, following analyses focus predominantly on the two equally trained DCNNs, as they are more suitable for a fair comparison of architectures.

Remarkably, vNet outperformed ResNet18 throughout all grouping variables (animacy, arousal, valence, and category) and generally seemed to be closer to the human benchmark level of accuracy (see Figure 4). However, across categories, applied Kruskal Wallis tests revealed that both ResNet18 ($X^2$(29) = 60.18, p < .001, r = .14) and vNet ($X^2$(33) = 58.65, p < .001, r = .11) performed significantly dissimilar to human observers and each other ($X^2$(33) = 62.76, p = .001, r = .61). Moreover, based on previous findings, we hypothesized that human observers should be significantly better at recognizing animate compared to inanimate objects (New, Cosmides, & Tooby, 2007). However, Wilcoxon rank sum tests indicated the existence of this effect but in the opposite direction, as human observers were significantly more accurate in recognizing inanimate (Median = 93.27) compared to animate objects (Median = 88.68; W = 345.5, p <.001, r = .55). Additionally, ResNet18 and vNet mirrored this behavior with a substantial increase in accuracy from animate (ResNet18: 61.90% / vNet: 68.10%) to inanimate objects (ResNet18: 79.05% / vNet: 84.76%).

Furthermore, low, medium, and high arousal and valence groups, obtained by tertile splits of human observer ratings (33rd and 66th percentile, see 4.2 Methods), revealed a significant, negative Spearman rank correlation between human accuracy and arousal (rho = -.22, p < .001). Interestingly, this relationship did not disappear when partial correlations controlling for animacy were computed (rho = -.20, p = .001). Subsequent Kruskal Wallis tests hinted at an effect of arousal ($X^2$(2) = 38.51, p < .001, r = .50) and Bonferroni-corrected, pairwise Wilcoxon tests revealed a significant decrease of accuracy between all three increasing levels of arousal. A significant positive correlation (rho = .34, p < .001), partial correlation controlling for animacy (rho = .37, p < .001), and an effect ($X^2$(2) = 36.90, p < .001, r = .48) with a significant increase of accuracy between low and medium levels in Bonferroni-corrected post-hoc tests were found for valence. Similarly, the accuracies of ResNet18 and vNet seemed to follow both effects. Images that were assessed as more calm and positive by human observers during rating trials lead to better recognition performance in human observers and DCNNs.

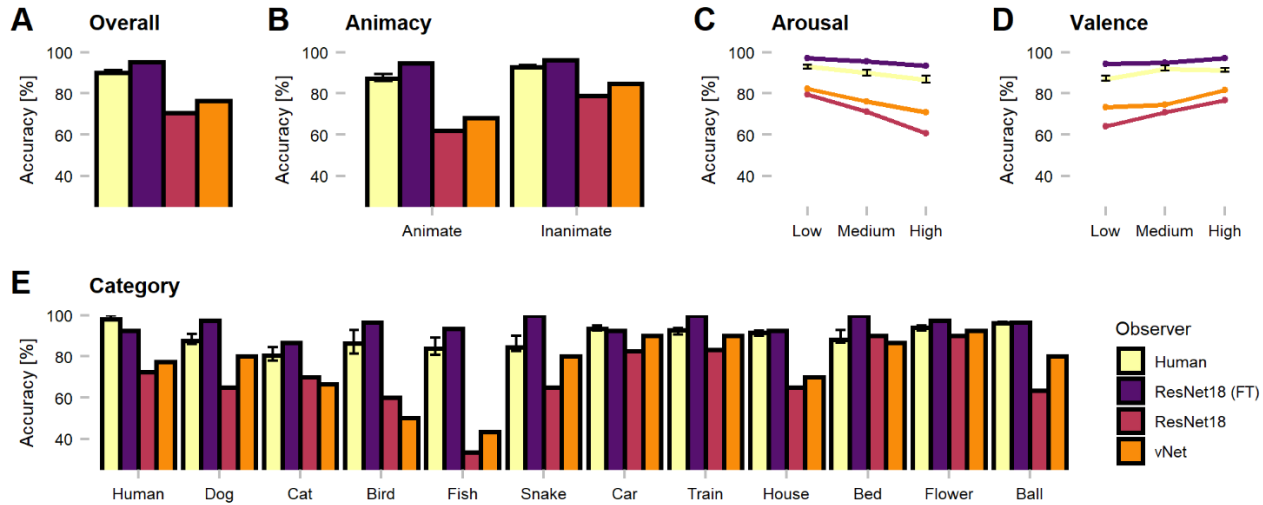

**Figure 4.** Object recognition accuracy of the human observer sample, fine-tuned ResNet18 (marked as FT), and both trained-from-scratch ResNet18 and vNet. Remarkably, vNet is more accurate then ResNet18 across all grouping variables. **(A)** Human observers were outperformed by fine-tuned ResNet18 but more accurate than both ResNet18 and vNet. **(B)** Human observers were better at recognizing inanimate compared to animate objects. This relationship held for DCNNs as well. **(C)** The effect of arousal on human observers indicated more inaccurate recognition with increasing arousal. **(D)** The effect of valence on human observers indicated more accurate recognition with increasing valence. **(E)** Human observers showed a small effect of category, while both trained from scratch DCNNs faced large variability between individual categories. **Note:** Confidence intervals for the human observer sample were estimated with Hodges-Lehmann procedure on a significance level of p = .05.

It is fundamental to note that only the performance of vNet exhibited a significant, positive Spearman rank correlation with human observer performance (see Table 1; rho = .14, p = .009), which may indicate a better fit to human categorization behavior. Contrary to our expectations, image properties (namely entropy, shape, texture, and power spectral peak-to-mean ratio) seemed to be rather unrelated to performance. Yet, as expected, the parameters were highly correlated among each other and demonstrated the statistical regularities of complex natural images.

**Table 1**

Spearman rank correlation between human observer accuracy, DCNN accuracy, and image properties.

| | Human Acc. | ResNet18 (FT) Acc. | ResNet18 Acc. | vNet Acc. | Entropy | Shape | Texture |
|---|---|---|---|---|---|---|---|
| Human Acc. | | | | | | | |
| ResNet18 (FT) Acc. | 0.071 | | | | | | |
| ResNet18 Acc. | 0.030 | 0.236*** | | | | | |
| vNet Acc. | 0.127** | 0.203*** | 0.551*** | | | | |
| Entropy | 0.041 | -0.071 | -0.053 | 0.027 | | | |
| Shape | 0.016 | -0.096* | -0.025 | -0.019 | 0.299*** | | |
| Texture | 0.036 | -0.056 | -0.044 | 0.000 | 0.432*** | 0.763*** | |
| Peak-to-Mean | 0.015 | 0.058 | 0.072 | 0.048 | -0.271*** | -0.591*** | -0.874*** |

**Note.** *Acc.* stands for Accuracy. *Peak-to-Mean* stands for Power Spectral Peak-to-Mean Ratio. Computed correlation used spearman-method with listwise-deletion. * = p < .05, ** = p < .01, *** = p < .001.

In order to shine more light on the classification errors made by human observers and DCNNs, which underlie the reported performances, categorization patterns were investigated (see Figure 5). Interestingly, as human observers seemed to have difficulties with relatively common classes (such as dog and cat or fish and snake), their classification behavior suggests that these conceptually similar classes could have lead to confusions. It should also be taken into account that other top-down and bottom-up influencing factors such as contextual cues may be especially similar between these classes. Moreover, both trained-from-scratch ResNet18 and vNet were found to misclassify images in similar ways.

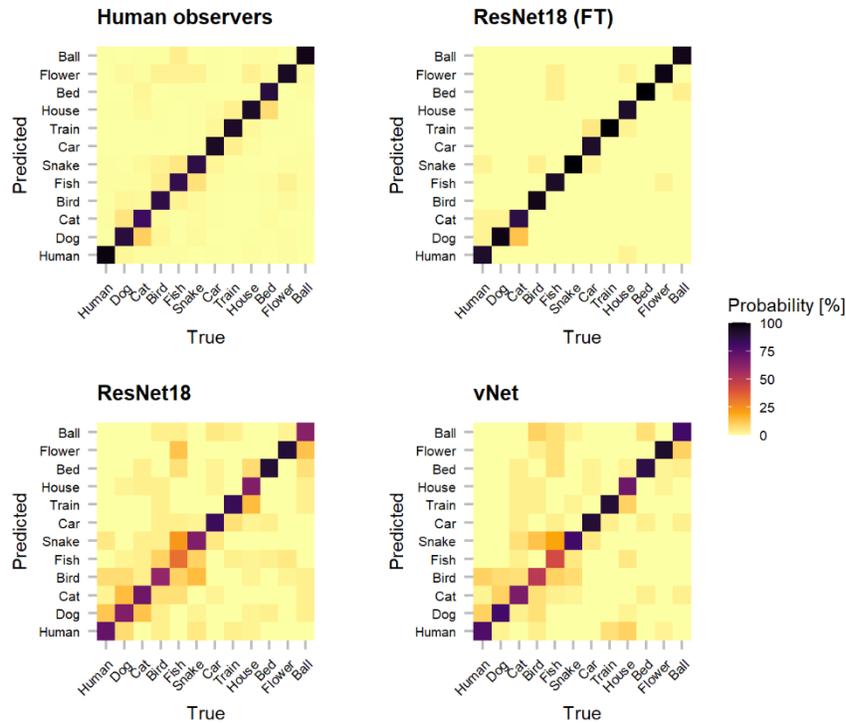

**Figure 5.** Categorization matrices reveal the classification patterns of true versus predicted labels that underly raw object recognition performances and thereby help to understand especially classification errors. The off-diagonal misclassifications show that human observers seemed to have problems with conceptually similar classes (such as dog and cat or fish and snake). Interestingly, especially both trained-from-scratch ResNet18 and vNet made similar mistakes.

## 3.2 Fixations – Feedforward versus Feedforward Processing

In an attempt to identify priorities during feedforward information processing in human observers and DCNNs, we inspected human fixations and global maxima of saliency maps. Fixations were compiled for the first 150ms, the theoretical time of a feedforward pass, and assigned to 1 out of 16 equally sized target blocks. Similarly, for GradCAM, the centroid of the single highest scoring patch was defined as the global maximum and used as an equivalent with the respective target block. As displayed in Figure 6, we found that human fixations were subject to a *central fixation bias* (Rothkegel, Trukenbrod, Schütt, Wichmann, & Engbert, 2017; Tatler, 2007), as most individual and almost all average fixations were located within the center blocks. Furthermore, a *saccadic motor bias* was visible, as average fixations of images presented on the left side were predominantly located on the right side of the image and vice versa. This pattern is thought to reflect the preference of the saccadic system for smaller amplitude eye movements over larger ones (Tatler, Baddeley, & Vincent, 2006). In most cases however, meaningful fixations on object features could be clearly identified on the individual level.

Although we hypothesized GradCAM maxima to differ substantially from human fixations, the results of ResNet18 and vNet across early, middle, and late layers proposed fundamentally diverging distributions deeper down the architectures. While human fixations were found to be normally distributed on a continuous scale of coordinates, GradCAM maxima, especially in middle and late layers, were located on discrete grids of different sizes. As these grids of possible maxima (ResNet18 Late = 7x7 and vNet Late = 4x4) were identical with the output sizes of the respective layers (see Figure 1, B + C), we attributed this behavior to both the agglomerative nature of convolutions and the resulting technical aspects of how the GradCAM algorithm extracts class activations from layers. To our knowledge, this characteristic of DCNN attribution methods has not been explicitly considered during previous human-machine comparisons so far. Since these findings restrict further comparisons of Euclidean distance measures between human fixations and GradCAM maxima, we proceeded by investigating only the specific target blocks, in which the respective fixations or maxima fell. As displayed in Figure 7, this analysis promoted more similar object recognition priorities between human observers and vNet.

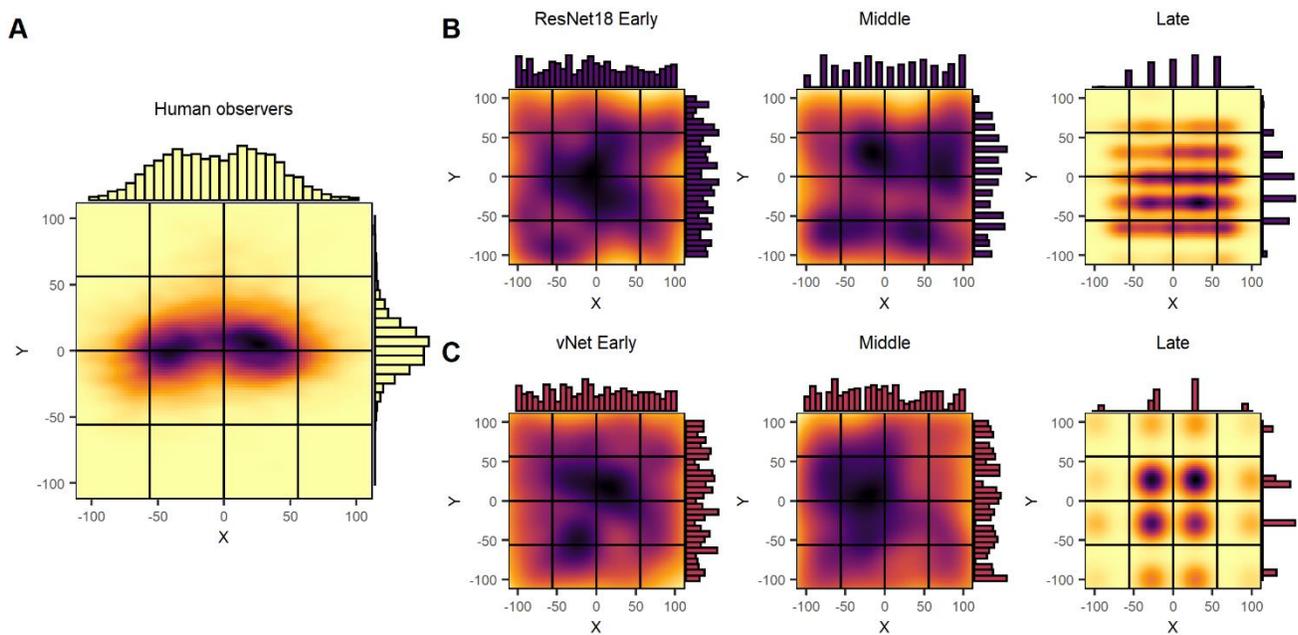

**Figure 6.** Uncorrected spatial priorities during feedforward object recognition in human observers and DCNNs. **(A)** Human fixations were affected by a central fixation and saccadic motor bias, while being normally distributed on a continuous scale. **(B + C)** ResNet18 and vNet GradCAM maxima displayed a near uniform distribution in early and middle layers with a 7x7 and 4x4 grid in the late layers. Generally, maxima followed a discrete segmentation that was identical to the output size of the respective layer.

In principle, the agreement between human observers and DCNNs in target blocks (see Figure 7) was rather low. The results indicated persisting differences after correcting for the discovered mismatches in resolution, as vNet coincided with human target blocks in 17.14% compared to ResNet18 with 11.67%. Generally, the agreement seemed to be higher for animate objects compared to inanimate objects and vNet prioritized especially more human target blocks on images of specific animate objects (especially *human* and *dog*). These findings further strengthened our confidence that vNet compared to ResNet18 indeed utilizes spatial priorities that are more similar to those of human observers during feedforward object recognition. These insights offer compelling evidence for a more human-like vNet and generally the impact of a RFS modification in terms of fit to human eye tracking data.

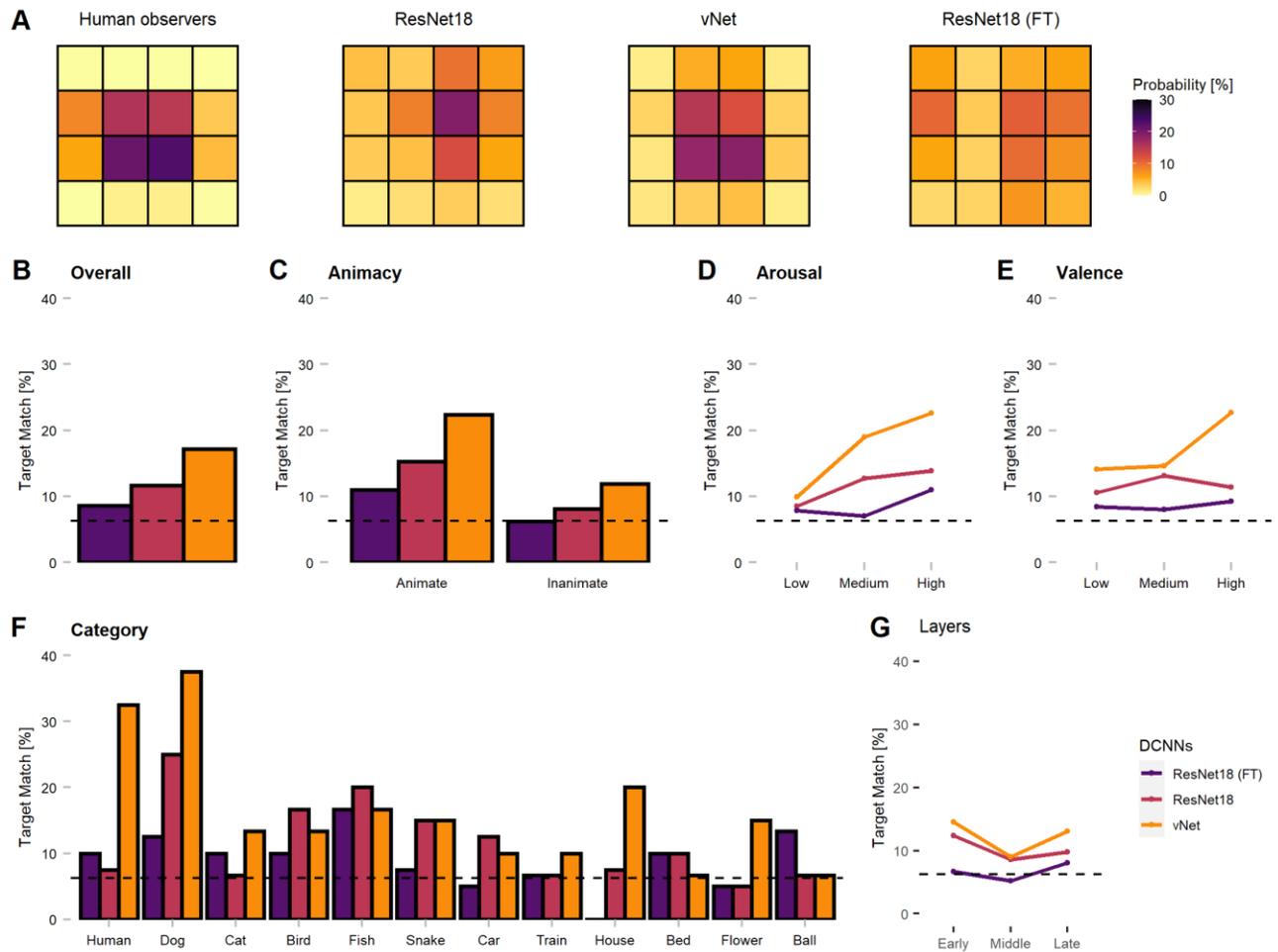

**Figure 7.** Corrected spatial priorities during feedforward object recognition in human observers and DCNNs' late layers. As the smallest resolution of output sizes allowed this resolution, all fixations and maxima were assigned to respective target blocks accordingly. **(A)** The target block percentages implied more similar spatial priorities between human observers and vNet, as mostly center blocks were targeted. In contrast, both ResNet18 models focused more on marginal target blocks. **(B)** Match between human and DCNN late layer target blocks across different grouping variables. vNet matched human target blocks more frequently. **(C - E)** DCNNs had a higher agreement on images of animate objects, with higher arousal ratings, and higher valence ratings. **(F)** vNet obtained especially high agreement on specific categories such as *human* and *dog*, while fine-tuned ResNet18 seemed to systematically choose different target block in *house* images. **(G)** Surprisingly, GradCAM maxima matched most frequently with human fixations in early, decreased in middle, and increased again in late layers. **Note:** The dotted line represents chance level at 6.25%.

## 3.3  Heatmaps – Recurrent vs. Feedforward Processing

Further analyses were conducted to compare human observers and DCNNs based on eye tracking and saliency heatmaps. Here, the focus was shifted away from feedforward mechanisms, as additionally recurrent processes with human eye movements after 150ms were investigated. We hypothesized that differences, which had already existed during early processing (i.e., effects of animacy, arousal, and valence), should be amplified in the brain due to mostly top-down processes setting in during this time window. It is important to note that this analysis is rather unfair in its nature, as it compares feedforward processing in DCNNs with additional recurrent processing in humans. However, in the light of this knowledge, it is entirely possible to test further hypotheses. Generally speaking, the obtained heatmaps illustrated the expected idiosyncrasies. With a few exceptions, human observers fixated the specific object within the first milliseconds and later shifted their attention to more relevant features (such as faces or arousing image parts). In contrast, DCNNs displayed their hierarchical organization with activation of especially specific shape and texture features in early, feature groups in middle, and whole objects in late layers (see Figure 8).

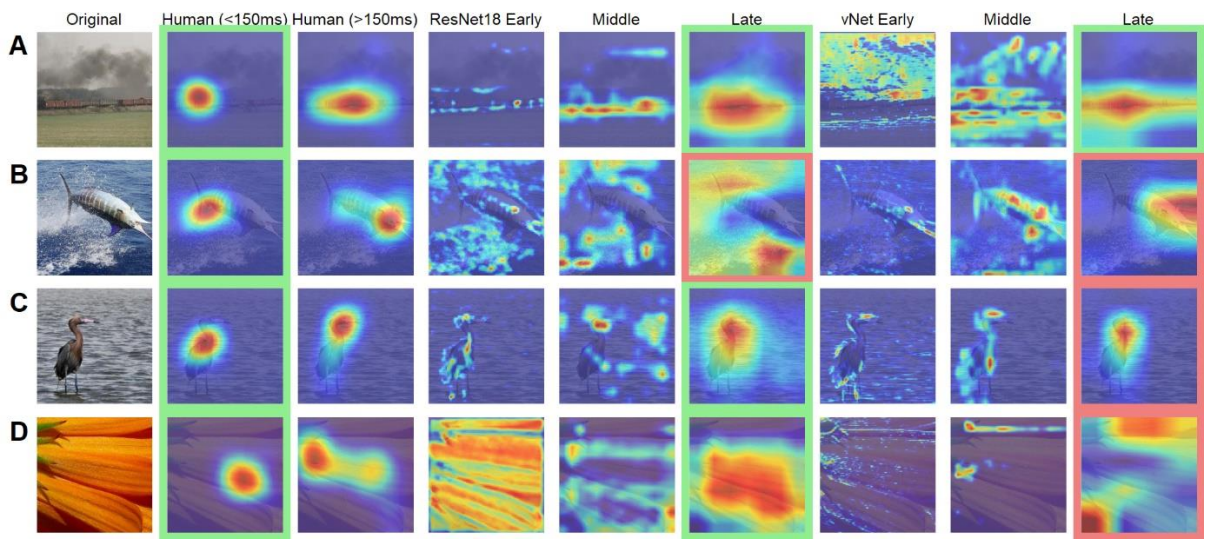

**Figure 8.** Examples of human observer heatmaps and DCNN GradCAMs of different layers. Images with **(A)** the highest and **(B)** the lowest correlation between human observers and ResNet18. Images with **(C)** the highest and **(D)** the lowest correlation between human observers and vNet. **Note:** Images highlighted in green were categorized correctly while images highlighted in red were categorized incorrectly.

Mean absolute error (MAE), defined as the individual deviation of a GradCAM from its human heatmap equivalent, was computed for all individual images. On average, MAEs of .29 for ResNet18, and .22 for vNet were found. These results fit well with previous findings by Ebrahimpour et al. (2019), who reported values of around .40 for a scene viewing task, and our previous outcomes promoting vNet as a better model for human eye tracking heatmaps. On top of that, MAE did not seem to be associated with general performance, as the fine-tuned ResNet18, which significantly outperformed human observers, reached only .32.

$$MAE = \frac{1}{Wx\,H} \sum_{x=1}^{W} \sum_{y=1}^{H} |E(x, y) - S(x, y)|$$

**Equation 1.** Mean absolute error (MAE) where W and H are the width and height of the original image in pixels (here 224x224), while E and S are the eye tracking heatmap and saliency map of the same image.

In order to link these results to the aforementioned control versus challenge distinction by Kar et al. (2019), we treated images which were categorized correctly by human observers (avg. accuracy = 100%) and DCNNs as control images (n = 145) and images which were categorized correctly by human observers (avg. accuracy = 100%) but categorized incorrectly by ResNet18 and vNet as challenge images (n = 60). Analyses across individual layers showed that the reported difference in MAE between both DCNNs seemed to emerge in late layers. These findings suggest that control and challenge images do not seem to be treated differently by the visual system in terms of their spatial priorities of information processing. Taken together with the findings of Kar et al. (2019) this means that the match between eye tracking and GradCAM data was not influenced by this distinction based on their difficulty.

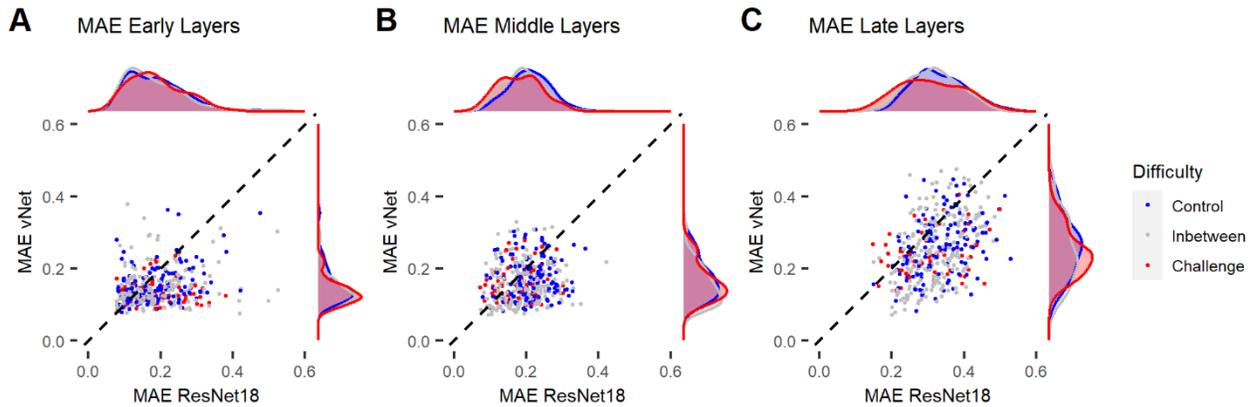

**Figure 9.** Mean absolute error (MAE) of ResNet18 and vNet from human observer heatmaps for individual images. Control images were categorized correctly by human observers, ResNet18, and vNet, while challenge images were categorized correctly by human observers but categorized incorrectly by ResNet18 and vNet. **(A + B)** MAEs seemed to be lowest in early layers. Here, the funnel shaped distribution hinted towards a lower boundary which may be a consequence of the discrete scale of GradCAMs. **(C)** As the distribution spread apart in later layers, the clearly smaller MAE of vNet became visible as more images could be found below the diagonal break-even line.

## 3.4 Arousal and Valence

On average, human observers rated images with median scores of 4.04 in arousal and 4.27 in valence (see Figure 10). In terms of arousal, Kruskal Wallis tests suggested that animate objects received significantly higher scores (Median = 4.41) compared to inanimate objects (Median = 3.59; W = 37971, p < .001, r = .28), whereas for valence, no significant difference was discovered (W = 22555, p = .685, r = .02). Here, a substantial disagreement is evident, as mean heart rate (arousal: $X^2(2) = 0.62$, p = .732, r = .02 / valence: $X^2(2) = 5.40$, p = .067, r = .05) and skin conductance response (arousal: M = X; $X^2(2) = 4.26$, p = .119, r = .04 / valence: M = X; $X^2(2) = 0.53$, p = .768, r = .03) did not differ substantially between images of low, medium, and high arousal and valence ratings. However, available ratings from the OASIS dataset (arousal: Median = 4.06 / valence: Median = 3.92) were more or less consistent with our ratings.

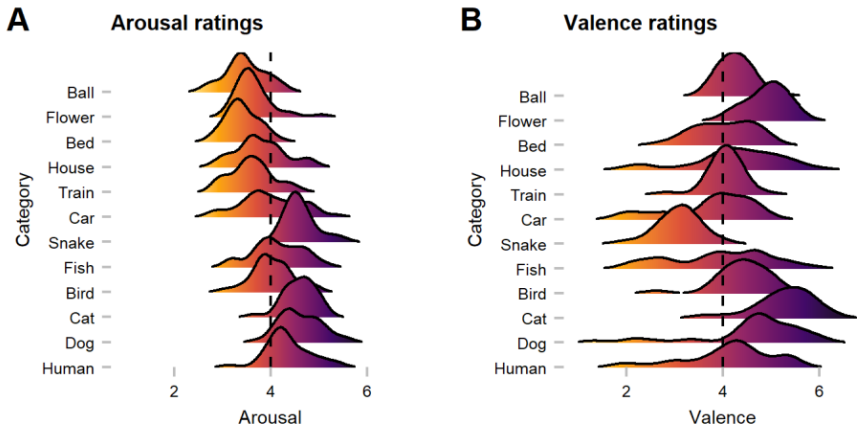

**Figure 10.** Human observers' arousal and valence ratings across categories. **(A)** Arousal ratings showed that animate categories were perceived as more arousing when compared to inanimate categories. **(B)** Valence ratings suggest category-specific effects (i.e., *snake*), as with a few exceptions most images were rated as rather positive. **Note:** The dotted line represents neutral scores.

# 4    Discussion

In this proof-of-concept study, we investigated the similarity of information processing in human observers and feedforward DCNN models during object recognition. For this purpose, human eye tracking heatmaps were compared to saliency maps of GradCAM, a customary attribution technique. Most importantly, during this endeavor, we found that GradCAM outputs, unlike eye tracking heatmaps, are produced on a discrete scale. While this is clear given the construction of DCNNs, to the best of our knowledge, this phenomenon has neither been regarded in previous studies nor stated explicitly in the literature of the field. As a natural consequence, this finding constrains several established results, as for example by Ebrahimpour et al. (2019), and also our own heatmap comparisons, as different resolutions underlying these visualizations pose an evident challenge. Therefore, it seems necessary to draw attention to this fact, as it might endanger the fairness of human-machine comparisons (Firestone, 2020; Funke et al., 2020). We even believe to find more evidence for this problem as displayed in Figure 9, where especially in earlier layers MAE values seemed to hit a lower boundary which was possibly due to fundamental mismatches in resolution and therefore impossible to undercut. Nevertheless, final comparisons of fixations and maxima were not influenced by this effect, as we controlled for spatial inaccuracy by proceeding with analyses on the target block level of the lowest maxima grid.

Generally, our results corroborate the assumption that the novel vNet architecture by Mehrer et al. (2021) captures human object recognition behavior more accurate compared to standard DCNNs commonly used throughout computer vision problems. Interestingly, this seemed to be the case on a performance level, as only vNet's performance was significantly correlated to human performance, and on a functional level of both before 150ms, as it matched target blocks of human fixations spatially more consistent, and after 150ms, where it yielded a lower MAE in late layers. On top of that, we argue that this higher similarity was not a side effect of higher performance compared to the equally trained ResNet18, as the fine-tuned ResNet18 model even outperformed human observers significantly and yet agreed the least with human fixations during feedforward processing. As already stated in previous literature (i.e., Dodge & Karam, 2017), the covariation in performance of human observers and DCNNs was found to be rather small. Nevertheless, these findings should be examined in the light of their relative impact, as vNet's correlation with human accuracy was not only of statistically significant importance, but also substantially higher. However, the reported effects should be treated with caution, as human-machine comparisons are prone to a wide range of confounding factors, either related to human cognition, such as the reported top-down and bottom-up processes, or related to deep learning problems, such as training settings and learning algorithms. We are aware of the fact that the number of parameters of both ResNet18 models and vNet is of necessity substantially different. In our view, these results emphasize the validity of a RFS modification as a method for designing more human-like models in computer vision. Moreover, as demonstrated by Luo, Li, Urtasun, and Zemel (2016), effective RFS follows a Gaussian distribution and become heavily increased by deep learning techniques such as subsampling (in most cases average or max pooling) and dilated convolutions, which are all commonly used in current architectures. Hence, as in contrast to DCNN saliency maps, human viewing behavior is highly focal, it would be interesting to see if the match of spatial priorities can be further increased in models that lack these computations. At the same time, future studies should target this topic by comparing DCNNs that only differ in RFS along their hierarchical architecture.

Moreover, we were able to identify control and challenge images based on the notion of Kar et al. (2019). Our analyses suggested no effect of image difficulty on the similarity between eye tracking heatmaps and saliency maps. While the authors' original findings on neural activity show that control and challenge images are processed differently especially during late time periods (>150ms), their results also propose that the two image groups share a similar early response and may not be treated differently by the visual system via the retina at all. Our reported null result regarding the viewing behavior may complement this line of argument well, as the authors even mentioned that on visual inspection no specific image properties differed between the groups. Unfortunately, in this case, the allocation to early- and late-solved images through neural recordings was not possible in the experimental setup at hand.

Surprisingly, in contradiction with earlier findings in humans (New et al., 2007), our results indicated that both human observers and DCNNs were more accurate in recognizing inanimate compared animate objects. This advantage for inanimate objects has been also reported on the level of basic categories by other studies before (Praß, Grimsen, König, & Fahle, 2013). Meanwhile, we discovered substantially more agreement between human and DCNN target blocks for animate objects. In turn, this suggests that both human observers and DCNNs were prioritizing more similar features during object recognition of animate objects. Therefore, we believe that this effect may be due to high efficient face processing mechanisms (Crouzet, Kirchner, & Thorpe, 2010), which could have led to more similarity in specific *face-heavy* categories (such as *human* and *dog*). Furthermore, as animacy could not fully explain arousal and valence effects on a behavioral and functional level of DCNNs, we argue that these effects could be a consequence of naturally learned optimization effects in visual systems, which have also been reported for image memorability judgments that automatically develop in DCNNs trained on object recognition and even predict variation in neural spiking activity (Jaegle et al., 2019; Rust & Mehrpour, 2020). This interpretation however needs to be treated with caution, as arousal and valence scores were obtained by human ratings which again underlie a wide range of effects such as for example acquired knowledge, attention, and memorability.

To summarize, in this paper we outline a novel concept of comparing human and computer vision during object recognition. In theory, this approach seems suitable for evaluating similarities and differences in priorities of information processing and may help to further pinpoint the specific impact of model adjustments towards more biological plausibility. We demonstrate this by showing that a RFS modification, which agrees conceptually with the ventral visual pathway, increases the model fit to human viewing behavior. Practically, we believe that our method will be improved by including different attribution techniques such as *Occlusion Sensitivity* (Zeiler & Fergus, 2014), which estimates class activations through a combination of occlusions and classifications, and thereby allow more adequate and comparable resolutions in the future. Furthermore, we hope that our idea will open up new perspectives on comparative vision at the intersection between biological and computer vision research.

# 5    Acknowledgements

We thank Michael Christian Leitner and Stefan Hawelka for their support regarding the eye tracking measurements, as well as Frank Wilhelm and Michael Liedlgruber for sharing their experience regarding the emotional stimuli and physiological measurements.